\documentclass[runningheads]{llncs}
\usepackage[T1]{fontenc}
\usepackage{graphicx}
\usepackage{amsmath,amssymb}
\usepackage{booktabs}
\usepackage{multirow}
\usepackage{xcolor}
\usepackage[hidelinks]{hyperref}
\usepackage{microtype}

\usepackage{xurl}

\newcommand{\model}{PRISM}
\newcommand{\best}[1]{\textbf{#1}}
\newcommand{\second}[1]{\underline{#1}}

\begin{document}

\title{PRISM: Exploring Heterogeneous Pretrained EEG Foundation Model Transfer to Clinical Differential Diagnosis}

\author{
Jeet Bandhu Lahiri\inst{1} \and
Parshva Runwal\inst{2} \and
Arvasu Kulkarni\inst{2} \and
Mahir Jain\inst{2} \and
Aditya Ray Mishra\inst{2} \and
Siddharth Panwar\inst{1,2} \and
Sandeep Singh\inst{2}
}

\authorrunning{Lahiri et al.}

\institute{
Indian Institute of Technology Mandi, India \\
\email{d23146@students.iitmandi.ac.in}
\and
NeuroDx, India \\
\email{parshva@neurodx.ai, arvasu@neurodx.ai, mahir@neurodx.ai, aditya.mishra@neurodx.ai, siddharth@neurodx.ai, sandeep@neurodx.ai}
}

\maketitle

\begingroup
\renewcommand\thefootnote{}
\footnotetext{Jeet Bandhu Lahiri, Parshva Runwal, and Arvasu Kulkarni contributed equally as first authors. Mahir Jain and Aditya Ray Misra contributed equally as second authors.}
\endgroup

\begin{abstract}
EEG foundation models are typically pretrained on narrow-source clinical archives and evaluated on benchmarks from the same ecosystem, leaving unclear whether representations encode neural physiology or recording-distribution artifacts. We introduce \model{} (\textbf{P}opulation-\textbf{R}epresentative \textbf{I}nvariant \textbf{S}ignal \textbf{M}odel), a masked autoencoder ablated along two axes---pretraining population and downstream adaptation---with architecture and preprocessing fixed. We compare a narrow-source EU/US corpus (TUH + PhysioNet) against a geographically diverse pool augmented with multi-center South Asian clinical recordings across multiple EEG systems. Three findings emerge. First, narrow-source pretraining yields stronger linear probes on distribution-matched benchmarks, while diverse pretraining produces more adaptable representations under fine-tuning---a trade-off invisible under single-protocol evaluation. Trained on three source corpora, \model{} matches or outperforms REVE (92 datasets, 60{,}000+ hours) on the majority of tasks, demonstrating that targeted diversity can substitute for indiscriminate scale and that dataset count is a confounding variable in model comparison. Second, on a clinically challenging and previously untested task---distinguishing epilepsy from diagnostic mimickers via interictal EEG---the diverse checkpoint outperforms the narrow-source checkpoint by +12.3 pp balanced accuracy, the largest gap across all evaluations. Third, systematic inconsistencies between EEG-Bench and EEG-FM-Bench reverse model rankings on identical datasets by up to 24 pp; we identify six concrete sources including split construction, checkpoint selection, segment length, and normalization, showing these factors compound non-additively.

\end{abstract}

\section{Introduction}

EEG foundation models---self-supervised models pretrained on scalp EEG---have produced strong transfer results across motor imagery, sleep staging, and abnormality detection~\cite{yang2024biot,jiang2024labram,wang2024cbramod,elouahidi2025reve}. Yet two fundamental questions remain unaddressed. First, \emph{what do these representations encode?} Pretraining corpora are drawn overwhelmingly from TUH, PhysioNet, and MOABB---repositories predominantly from Europe and North America where recording conditions, amplifier characteristics, and population physiology are confounded within a narrow distribution~\cite{elouahidi2025reve}. A model that achieves low reconstruction error on such data may have memorized recording-environment regularities rather than neural dynamics, and standard benchmarks from the same ecosystem cannot distinguish these explanations. Second, \emph{can we trust existing benchmark rankings?} EEG-Bench~\cite{kastrati2025eegbench} and EEG-FM-Bench~\cite{xiong2025eegfmbench} are the two most widely used standardized evaluation frameworks for EEG foundation models, yet they make different methodological choices in ways that produce not merely quantitative discrepancies but outright ranking reversals on the same models and datasets.

We investigate both questions through controlled ablations with \model{}, a masked autoencoder inspired from the REVE architecture~\cite{elouahidi2025reve}. Three findings emerge.

First, population heterogeneity does not uniformly improve downstream accuracy---it changes what representations are \emph{good for}. We construct a narrow-source pool (TUH~+~PhysioNet) and a geographically diverse pool adding multi-center South Asian clinical recordings. Under linear probing, the narrow-source checkpoint leads on distribution-matched benchmarks; under fine-tuning, the diverse checkpoint matches or exceeds it on most tasks. Notably, \model{} trained on three source corpora matches or outperforms REVE trained on 92 datasets, demonstrating that targeted data diversity can substitute for indiscriminate data scale---and that dataset count is a confounding variable in EEG foundation model comparison.

Second, we provide a concrete analysis of six specific methodological differences between EEG-Bench and EEG-FM-Bench that produce divergent results: train/validation split construction, checkpoint selection strategy, input segment length, preprocessing normalization, classification head selection, and self-reported versus standardized evaluation. We show that correcting individual factors resolves discrepancies for some models but not others, revealing compounding interactions. These inconsistencies underscore the urgent need for a community consensus on evaluation methodology---without which model comparisons are unreliable.

Third, we introduce the first evaluation of an EEG foundation model on distinguishing epilepsy from its diagnostic mimickers---including psychogenic non-epileptic seizures (PNES), syncope, and other paroxysmal disorders---using routine interictal EEG, a clinically relevant task that has not previously been addressed by any foundation model. This task is substantially harder than existing benchmarks: up to 25\% of patients with refractory epilepsy diagnoses are later found to have been misdiagnosed~\cite{benbadis2009differential}, and the mean diagnostic delay for PNES alone exceeds five years~\cite{reuber2002diagnostic}. On this task, the diverse checkpoint significantly outperforms the narrow-source checkpoint, providing direct evidence that pretraining population composition impacts clinically meaningful downstream performance.

\section{Related Work}

\textbf{EEG foundation models.} Self-supervised pretraining for EEG has advanced through BIOT~\cite{yang2024biot} (cross-data transformer on TUH), LaBraM~\cite{jiang2024labram} (vector-quantized codebook across TUH and MOABB), CBraMod~\cite{wang2024cbramod} (criss-cross attention), and REVE~\cite{elouahidi2025reve} (92 datasets, 25{,}000 subjects, 4D Fourier positional encoding enabling transfer across arbitrary montages---an architectural innovation we adopt). All share a limitation: pretraining data originates predominantly from EU/US repositories, and no prior work has tested whether broadening the population changes \emph{what} the model learns.

\textbf{Pretraining data scale as a confounding variable.} A prevailing assumption in EEG foundation model development is that larger pretraining corpora yield better downstream representations, mirroring scaling intuitions from language modeling. This has driven a progression from single-corpus models trained on TUH alone to REVE's corpus spanning 92 datasets, 25{,}000 subjects, and over 60{,}000 hours. However, dataset \emph{quantity} and representation \emph{quality} are not equivalent, and the EEG domain has not validated that scaling laws from language transfer to neural signals. EEG-FM-Bench itself observes that ``scaling data often fails to yield proportional gains in downstream tasks''~\cite{xiong2025eegfmbench}, and empirical work on EEG pathology classification has found ceiling effects when scaling labeled data beyond moderate corpus sizes~\cite{kiessner2024ceiling}. Critically, when models differ simultaneously in architecture, pretraining objective, corpus size, and corpus composition, it is impossible to attribute performance differences to any single factor. No prior work has performed controlled dataset ablations---varying corpus composition while holding architecture, objective, and evaluation protocol fixed---to establish what data actually contributes. This confound renders published dataset-count comparisons between models largely uninterpretable as evidence of model quality.

\textbf{Evaluation inconsistencies.} EEG-Bench~\cite{kastrati2025eegbench} and EEG-FM-Bench~\cite{xiong2025eegfmbench} each provide standardized evaluation protocols but make different choices in preprocessing, segmentation, checkpoint selection, and classifier heads. As we show in Section~\ref{sec:benchmark}, these choices can reverse model rankings on the same dataset, with discrepancies exceeding 24 percentage points. This mirrors findings in medical imaging, where evaluation methodology determines whether a model appears to generalize~\cite{zech2018variable}. We provide the first systematic decomposition of the specific sources of these inconsistencies and their interactions.

\textbf{Demographic bias in medical AI.} Chest X-ray models exhibit systematic underdiagnosis of marginalized subgroups~\cite{gichoya2024bias}, and most clinical AI algorithms train on data from only three geographic regions~\cite{kaushal2020geographic}. These failures arise from distributional shortcuts---models encode hospital site rather than pathology~\cite{zech2018variable,degrave2021ai}. The analogous concern for EEG has not been tested until this work.

\textbf{Epilepsy vs.\ mimickers.} Distinguishing epilepsy from its mimickers is among the hardest problems in clinical neurophysiology. PNES accounts for 20--30\% of epilepsy monitoring unit admissions~\cite{benbadis2007pnes}, with diagnostic delays exceeding five years~\cite{reuber2002diagnostic} and misdiagnosis rates of up to 25\%~\cite{benbadis2009differential}. While prior ML work addresses seizure detection~\cite{shoeibi2021seizure}, no study has applied EEG foundation models to differential diagnosis from interictal recordings.

\section{Method}

Our experimental design isolates the effect of pretraining population on downstream representation quality, evaluating across both standard benchmarks and a novel clinical task.

\subsection{Pretraining Pools}

We construct two dataset pools with identical preprocessing (200\,Hz resampling, 0.5--99.5\,Hz bandpass, notch filtering at 50 and 100\,Hz, per-channel Z-score normalization, $\pm$15$\sigma$ clipping):

\textbf{D1 (Narrow-source):} TUH Corpus~\cite{obeid2016tuh} + PhysioNet Motor Imagery~\cite{goldberger2000physiobank}---the standard pretraining distribution used by most EEG foundation models, predominantly from North American and European recording centers under uniform clinical protocols.

\textbf{D2 (Multi-source):} D1 augmented with multi-center clinical recordings from 9663 subjects totalling 4170 hours from South Asian institutions, introducing geographic, demographic, and acquisition-system heterogeneity---different EEG systems (including Natus, Nihon Kohden, and BPL units), referencing conventions, and electrode impedance profiles.

No balancing or re-weighting is applied. All recordings are mapped to the standard 10--20 montage (19 channels) via automated channel-name normalization handling common aliases (T7$\to$T3, P7$\to$T5) and acquisition-specific prefixes.

\subsection{Architecture}

\model{} adopts the masked autoencoder architecture introduced by REVE's 4D positional encoding~\cite{elouahidi2025reve}, with modifications to the auxiliary training path. Each EEG channel $c$ with signal $\mathbf{x}_c \in \mathbb{R}^{T}$ is segmented into patches of $P{=}200$ samples (1\,s at 200\,Hz) with overlap $O{=}20$ samples (step size $S{=}P{-}O{=}180$), and each patch $\mathbf{p} \in \mathbb{R}^P$ is embedded via a convolutional projection $\mathbf{e} = \mathbf{W}_e \, \mathbf{p}$ where $\mathbf{W}_e \in \mathbb{R}^{D \times P}$ with $D{=}512$.

\textbf{4D Positional Encoding.} Following REVE~\cite{elouahidi2025reve}, each token receives a positional encoding from its electrode's spatial coordinates $(x,y,z)$ and temporal patch index $t$. Let $\mathbf{c} = [x, y, z, t]^\top \in \mathbb{R}^4$ denote the coordinate vector. We construct a frequency matrix $\mathbf{F} \in \mathbb{R}^{4 \times K}$ from the Cartesian product of $n_\text{freq}{=}4$ linearly spaced frequencies per dimension, yielding $K{=}n_\text{freq}^4{=}256$ combinations:
\begin{equation}
\text{PE}(\mathbf{c}) = \text{LN}\!\left(\mathbf{W}_f\left[\sin(\mathbf{F}^\top\mathbf{c});\;\cos(\mathbf{F}^\top\mathbf{c})\right] + \text{MLP}(\mathbf{c})\right)
\end{equation}
where $\mathbf{W}_f \in \mathbb{R}^{D \times 2K}$ is a learned projection and LN is layer normalization. This encoding enables processing of arbitrary electrode montages without retraining.

\textbf{Encoder--Decoder.} The encoder is a 12-layer pre-norm transformer (8 heads, $D{=}512$, GELU feedforward with expansion factor 4) processing only unmasked tokens. The decoder (4 layers, same configuration) receives encoder outputs at visible positions and learnable mask tokens at masked positions, reconstructing masked patches via a linear head.

\textbf{Masking.} Spatio-temporal block masking at ratio $\rho{=}0.55$ selects random seed tokens and masks all tokens within a 3\,cm spatial and 3\,s temporal radius, iterating until $\lfloor\rho N\rfloor$ tokens are masked; excess masked tokens are randomly restored to enforce exact batch-uniform counts.

\subsection{Training Objective}

The model is trained with two L1 reconstruction losses. Let $\hat{\mathbf{x}}_i$ and $\mathbf{x}_i$ denote the reconstructed and original signal for masked patch $i$, and let $\mathcal{M}$ denote the set of masked indices. The primary loss measures decoder reconstruction quality:
\begin{equation}
\mathcal{L}_\text{pri} = \frac{1}{|\mathcal{M}|}\sum_{i \in \mathcal{M}} \left\|\hat{\mathbf{x}}_i - \mathbf{x}_i\right\|_1
\end{equation}

An auxiliary path concatenates feedforward outputs from all encoder layers per token and pools them via a learned attention mechanism with a single query vector to produce a global embedding, which independently reconstructs each masked patch conditioned on the masked position's positional encoding. The auxiliary loss $\mathcal{L}_\text{sec}$ follows the same L1 form; by forcing reconstruction from a single global token, this path distributes information across encoder depth. The total loss is:
\begin{equation}
\mathcal{L} = \mathcal{L}_\text{pri} + \lambda \, \mathcal{L}_\text{sec}, \qquad \lambda = 0.1
\end{equation}

Two checkpoints are trained: D1 (narrow-source) and D2 (multi-source), both with identical architecture and training hyperparameters.

\begin{figure}[t]
\centering
\includegraphics[width=\linewidth]{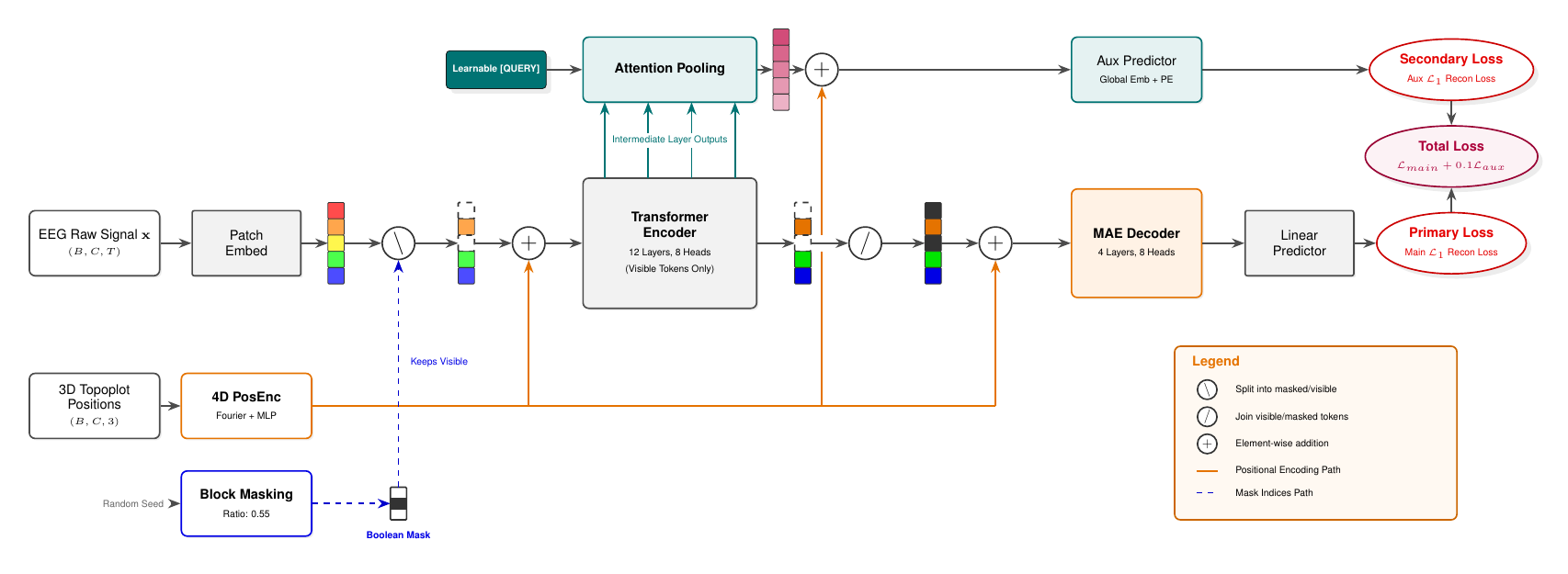}
\caption{\model{} architecture. Raw EEG is patchified per channel and embedded. 4D positional encoding~\cite{elouahidi2025reve} (electrode coordinates + temporal index) is added before the encoder. The decoder fills masked positions with learnable tokens and reconstructs signal. An auxiliary path pools intermediate encoder outputs via learned attention, producing a global embedding that independently reconstructs masked patches.}\label{fig:arch}
\end{figure}

\subsection{Downstream Tasks and Evaluation}

\textbf{Benchmark tasks.} We evaluate on six established tasks: ADFTD~\cite{miltiadous2023adftd} (Alzheimer's/ frontotemporal dementia, 2-class), BCI-IV-2a~\cite{tangermann2012bci} (motor imagery, 4-class), HMC~\cite{alvarezestevez2021hmc} (sleep staging, 5-class), PhysioNet-MI~\cite{goldberger2000physiobank} (motor imagery, 4-class), Siena Scalp~\cite{detti2020siena} (pathology, 2-class), and EEGMAT~\cite{zyma2019mental} (cognitive load, 2-class).

\textbf{Clinical task: Epilepsy vs.\ Mimickers.} We introduce a novel downstream task addressing a critical unmet clinical need. Distinguishing epilepsy from its diagnostic mimickers---primarily psychogenic non-epileptic seizures (PNES), vasovagal syncope, and other paroxysmal non-epileptic events---from routine interictal EEG represents one of the most diagnostically challenging problems in clinical neurophysiology. Unlike seizure detection tasks that operate on ictal recordings (which contain overt electrographic correlates), interictal EEG in this differential diagnosis problem is subtle: background activity may appear broadly similar across groups, and discriminative markers are sparse, non-specific, and highly dependent on careful reading. This setting has received no prior attention from EEG foundation model research, making it an important and underexplored benchmark for clinically meaningful evaluation.

The dataset comprises 200 subjects---100 with confirmed epilepsy and 100 with confirmed mimicker diagnoses---recorded on clinical-grade Natus EEG systems at 256\,Hz in the standard 10--20 montage at tertiary care clinical centers in India. All subjects are from the South Asian population, making this dataset geographically and demographically distinct from existing EEG benchmarks. Diagnosis was established by board-certified epileptologists following established criteria~\cite{lafrance2020pnes}, using the full clinical workup including video-EEG monitoring, ictal semiology assessment, longitudinal follow-up, and response to treatment. Cases where diagnosis remained uncertain at follow-up were excluded, ensuring label reliability.

We use subject-level splits: 85 subjects per class for training, 5 for validation, and 10 for held-out evaluation. Data is resampled to 200\,Hz and preprocessed identically to the pretraining pipeline. The dataset will be released as a public benchmark upon acceptance, constituting the first publicly available interictal EEG dataset curated for this differential diagnosis task; details regarding data access will be made available via the project repository at publication.

\textbf{Adaptation strategies.} We evaluate four regimes. \emph{Linear probing (LP)}: encoder frozen, only the classification head is trained. \emph{Full fine-tuning, single-stage (Full-Single)}: all parameters updated end-to-end from the pretrained checkpoint. \emph{Full fine-tuning, dual-stage (Full-Dual)}: first train the classification head with frozen encoder (LP), then unfreeze all parameters for joint training---preserving pretrained structure during initial adaptation. \emph{Partial fine-tuning, single-stage (Partial-Single)}: only the final $k$ encoder layers and classification head are updated, keeping early layers frozen. Classification head ablation is performed across attention pooling, average pooling, and a two-layer MLP. All results report balanced accuracy.

\section{Experiments and Results}

\subsection{Study 1: Effect of Population Composition}\label{sec:study1}

\begin{table}[t]
\caption{Effect of pretraining population on balanced accuracy across six benchmark tasks. LP = linear probing (frozen encoder, MLP head), FT = full fine-tuning single-stage with MLP head. \best{Bold} = best, \second{underline} = second best per section-column.}\label{tab:study1}
\centering
\scalebox{0.82}{
\begin{tabular}{llcccccc}
\toprule
& Ckpt & ADFTD & BCI-IV-2a & HMC & PhysioNet & Siena & EEGMAT \\
\midrule
\multirow{2}{*}{LP}
& D1 & \best{0.554} & \second{0.481} & \best{0.682} & \best{0.604} & \best{0.650} & \second{0.576} \\
& D2 & \second{0.533} & \best{0.507} & \second{0.679} & \second{0.593} & \second{0.638} & \best{0.649} \\
\midrule
\multirow{2}{*}{FT}
& D1 & \second{0.524} & \second{0.542} & \best{0.703} & \second{0.631} & \best{0.738} & \second{0.592} \\
& D2 & \best{0.545} & \best{0.543} & \second{0.673} & \best{0.640} & \best{0.738} & \best{0.648} \\
\bottomrule
\end{tabular}
}
\end{table}

Table~\ref{tab:study1} reveals a dissociation invisible when a single evaluation regime is reported. Under LP, D1 outperforms D2 on four of six tasks, consistent with D1's representations being aligned with evaluation data statistics. Under FT, D2 matches or exceeds D1 on five of six tasks. Diverse pretraining produces representations in a more general embedding region that requires nonlinear adaptation but ultimately yields superior performance. This is consistent with the evaluation instability documented by EEG-Bench and EEG-FM-Bench, where different pooling strategies and fine-tuning regimes routinely reverse model rankings~\cite{kastrati2025eegbench,xiong2025eegfmbench}.

The exception---HMC sleep staging, where D1 retains its FT advantage---is consistent with the neurophysiological distinctiveness of sleep macrostructure, defined by gross spectral features (delta dominance in N3, spindles in N2~\cite{berry2017aasm}) largely invariant to population and recording system.

\subsection{Study 2: Classification Head Ablation}\label{sec:study2}

\begin{table}[t]
\caption{Head ablation on D2, full fine-tuning. \best{Bold} = best, \second{underline} = second best per column.}\label{tab:head}
\centering
\scalebox{0.82}{
\begin{tabular}{lcccccc}
\toprule
Head & ADFTD & BCI-IV-2a & HMC & PhysioNet & Siena & EEGMAT \\
\midrule
Attn Pool & 0.485 & \second{0.410} & \best{0.707} & 0.604 & \second{0.737} & \second{0.636} \\
Avg Pool  & \second{0.527} & 0.389 & \second{0.705} & \second{0.627} & \best{0.787} & 0.607 \\
MLP       & \best{0.545} & \best{0.543} & 0.673 & \best{0.640} & 0.738 & \best{0.648} \\
\bottomrule
\end{tabular}
}
\end{table}

Table~\ref{tab:head} shows that the MLP head outperforms both pooling alternatives on four of six tasks within our model, with the largest margin on BCI-IV-2a ($+$13.3\,pp over attention pooling, $+$15.4\,pp over average pooling). The exceptions---HMC and Siena---are tasks where class boundaries align with coarse global features that simple aggregation captures adequately.

We hypothesize this reflects an interaction between the reconstruction pretraining objective and the aggregation mechanism. MAE tokens are optimized for patch-level reconstruction, encoding local waveform morphology rather than globally discriminative features. Attention pooling may inherit biases toward reconstruction-informative tokens rather than discriminative ones, while average pooling dilutes sparse discriminative content. The MLP applies a learned nonlinear re-projection from reconstruction space toward classification. EEG-FM-Bench reports analogous sensitivity to head choice across models~\cite{xiong2025eegfmbench}, suggesting the effect is not unique to our architecture.

\subsection{Study 3: Fine-Tuning Strategy}\label{sec:study3}

\begin{table}[t]
\caption{Adaptation strategy comparison on D2, MLP head. Full-Single = all parameters trained end-to-end; Full-Dual = LP stage then full fine-tuning; Partial-Single = only last encoder layer unfrozen. \best{Bold} = best, \second{underline} = second best per column.}\label{tab:ft}
\centering
\scalebox{0.82}{
\begin{tabular}{lcccccc}
\toprule
Strategy & ADFTD & BCI-IV-2a & HMC & PhysioNet & Siena & EEGMAT \\
\midrule
LP              & 0.533 & 0.507 & 0.679 & 0.593 & 0.638 & \second{0.649} \\
Full-Dual       & \best{0.557} & \second{0.521} & \best{0.696} & 0.598 & 0.613 & 0.635 \\
Full-Single     & \second{0.545} & \best{0.543} & 0.673 & \best{0.640} & \best{0.738} & 0.648 \\
Partial-Single  & 0.531 & 0.518 & \second{0.682} & \second{0.607} & \second{0.663} & \best{0.658} \\
\bottomrule
\end{tabular}
}
\end{table}

No strategy dominates (Table~\ref{tab:ft}). Full-Dual wins on ADFTD and HMC---tasks requiring distributed temporal characterization---where the two-stage approach preserves pretrained structure. Full-Single excels on motor imagery and Siena, where end-to-end feature reorganization around localized discriminative patterns is effective. Partial-Single achieves the best result on EEGMAT, balancing adaptation with preservation of early feature extraction layers. Adaptation strategy is a consequential experimental choice, not merely a hyperparameter.

\subsection{Study 4: Epilepsy vs.\ Mimickers}\label{sec:epilepsy}

\begin{table}[t]
\caption{Epilepsy vs.\ mimickers: balanced accuracy on held-out subjects. D1 = narrow-source checkpoint, D2 = multi-source checkpoint. Full-Single fine-tuning with MLP head. \best{Bold} = best, \second{underline} = second best.}\label{tab:epilepsy}
\centering
\scalebox{0.82}{
\begin{tabular}{lcc}
\toprule
 & D1 & D2 \\
\midrule
Full-Single & \second{0.484} & \best{0.607} \\
\bottomrule
\end{tabular}
}
\end{table}

Table~\ref{tab:epilepsy} presents results on the most diagnostically challenging task in our evaluation suite. The multi-source checkpoint (D2) outperforms the narrow-source checkpoint (D1) by 12.3\,pp in balanced accuracy---substantially larger than the gap on any benchmark task (Table~\ref{tab:study1}), where D1 and D2 are typically separated by fewer than 2\,pp under FT.

Epilepsy vs.\ mimicker discrimination from interictal EEG is a task that even expert neurologists find difficult---routine EEG sensitivity for epilepsy is $\sim$50\% in isolation~\cite{smith2005eeg}, and EEG overinterpretation is a well-documented source of misdiagnosis~\cite{benbadis2007errors}. That pretraining population produces a measurable 12.3\,pp difference on this task, when the evaluation data itself comes from a single center with uniform protocols, suggests the multi-source checkpoint has learned more robust neural representations that better capture subtle pathological distinctions. We hypothesize that multi-source pretraining forces the encoder to disentangle neural content from acquisition artifacts during reconstruction---an implicit invariance that the narrow-source encoder lacks.

Notably, the evaluation cohort is drawn exclusively from South Asian clinical centers, demographic and geographic territory well-represented in D2 but absent in D1. This demographic alignment between pretraining diversity and evaluation population may contribute to D2's advantage. Future work should disentangle the relative contributions of geographic representation and acquisition-system diversity.

\subsection{Study 5: Comparison with Existing Models and Cross-Protocol Inconsistency}\label{sec:benchmark}

\begin{table}[t]
\caption{Comparison with EEG foundation models under two standardized evaluation protocols on motor imagery tasks. Protocols differ in preprocessing, segmentation, checkpoint selection, and evaluation heads---see Section~\ref{sec:benchmark} for a systematic decomposition. \best{Bold} = best, \second{underline} = second best per column.}\label{tab:bench}
\centering
\scalebox{0.82}{
\begin{tabular}{lcccc}
\toprule
& \multicolumn{2}{c}{BCI-IV-2a} & \multicolumn{2}{c}{PhysioNet-MI} \\
\cmidrule(lr){2-3} \cmidrule(lr){4-5}
Model & EEG-Bench & FM-Bench & EEG-Bench & FM-Bench \\
\midrule
LaBraM      & 0.28 & \second{0.38} & 0.25 & \second{0.49} \\
CBraMod     & \second{0.31} & 0.35 & \second{0.27} & 0.46 \\
REVE        & \second{0.41} & 0.33 & \best{0.33} & 0.52 \\
\model{}    & \best{0.44} & \best{0.51} & \second{0.32} & \best{0.59} \\
\bottomrule
\end{tabular}
}
\end{table}

\model{} achieves the highest accuracy on three of four protocol--task combinations (Table~\ref{tab:bench}). However, the cross-protocol inconsistencies are at least as informative as the absolute rankings.

\textbf{Decomposition of protocol differences.} EEG-Bench and EEG-FM-Bench differ in six concrete methodological dimensions, each of which can independently shift results:

\textbf{(1) Train/validation split construction.} EEG-Bench uses a subject-level held-out test set, with the remaining data split randomly at the segment level (20\% validation), meaning some subjects' segments span both train and validation sets. EEG-FM-Bench maintains strict subject-level separation for all splits. This discrepancy means EEG-Bench's effective training set is equivalent to EEG-FM-Bench's combined train and validation sets, inflating EEG-Bench performance estimates for subject-independent generalization.

\textbf{(2) Checkpoint selection policy.} EEG-Bench selects the checkpoint achieving best validation performance (early stopping), while EEG-FM-Bench returns the last training checkpoint. For models prone to overfitting on small fine-tuning sets, this difference can produce large performance swings independent of model quality.

\textbf{(3) Input segment length.} EEG-FM-Bench processes 4-second windows whereas EEG-Bench uses 3-second windows by default. The additional second provides more temporal context at inference, systematically favoring models that capture longer-range temporal dependencies---a bias that interacts nontrivially with architectures that use windowed attention.

\textbf{(4) Preprocessing normalization.} The two frameworks apply different normalization schemes in their feature extraction pipelines. Normalization differences affect the effective amplitude distribution presented to the model, which can disproportionately affect models that were pretrained under specific normalization assumptions.

\textbf{(5) Classification head and fine-tuning methodology.} Self-reported metrics in individual model papers are often produced under the protocol most favorable to that model (e.g., attention pooling rather than MLP heads; specific learning rate schedules). EEG-FM-Bench imposes a fixed evaluation protocol across models, which is more comparable but may disadvantage models whose optimal heads were not included. Our Study~2 (Section~\ref{sec:study2}) demonstrates that head choice alone can shift performance by over 15\,pp within the same model on the same task.

\textbf{(6) Compound interactions.} We find that correcting for the validation split and segment length equalizes results for CBraMod on BCI-IV-2a, but analogous corrections do not resolve discrepancies for other models. This indicates that the six factors compound non-additively: a model optimized to exploit one protocol's structure may be systematically disadvantaged by another's, making it impossible to attribute ranking differences to any single methodological choice.

The comparison with REVE is instructive: both share similar architectural components but differ in pretraining composition and scale. REVE uses 92 datasets and 25{,}000 subjects from predominantly EU/US sources; \model{} uses three source corpora. On BCI-IV-2a, \model{} outperforms REVE under both protocols. On PhysioNet-MI, rankings reverse between protocols (REVE $+$1\,pp EEG-Bench; \model{} $+$7\,pp FM-Bench), illustrating that multi-source pretraining provides complementary value that is protocol-dependent in its expression.

\textbf{Dataset quantity as a confound.} Beyond protocol inconsistency, this comparison surfaces a structural problem in EEG foundation model evaluation: dataset count is routinely used as a proxy for model quality, yet it is a confounding variable whenever models also differ in architecture, pretraining objective, and corpus composition---as all published model pairs do. \model{} matches or exceeds REVE on three of four protocol--task combinations despite a roughly 30-fold difference in the number of source datasets. This does not imply that data quantity is unimportant; it implies that the marginal return of aggregating additional EU/US datasets past some saturation point may be limited relative to the return of \emph{targeted diversity}. Without controlled dataset ablations---varying corpus size while holding all other factors fixed---one cannot determine whether REVE's broader corpus or its other design choices drive its performance on tasks where it leads. The field therefore risks two compounding errors: (i) misattributing performance to data scale when the driving factor may be architecture or objective, and (ii) over-collecting data from redundant distributions that do not improve downstream representations. Systematic dataset ablations, in the spirit of our D1-vs-D2 comparison, should be a required component of any EEG foundation model publication. These findings collectively underscore that \emph{how} a model is evaluated and \emph{what} data it is trained on can each matter as much as the model itself, and that current reporting practices make it difficult to disentangle these factors~\cite{kastrati2025eegbench,xiong2025eegfmbench}.

\section{Discussion}

\textbf{The distribution-matching trap.} LP accuracy on narrow-source benchmarks systematically favors narrow-source pretrained models because their representations are pre-aligned with the evaluation distribution. The reversal under FT shows the diverse model encodes equivalent or superior information requiring adaptation to extract. For clinical systems that always undergo task-specific adaptation, LP-only evaluation is misleading.

\textbf{From benchmarks to clinical impact.} Pretraining population composition produces its largest effect on the clinically hardest task---epilepsy vs.\ mimickers. On standard benchmarks, D1 and D2 are often within 1--2\,pp under FT; on the clinical task, the gap widens to 12.3\,pp. Benchmark tasks---dominated by gross physiological distinctions---are insufficient proxies for clinical utility. The community would benefit from incorporating challenging differential diagnosis tasks into evaluation suites, and our released dataset is a step toward this goal.

\textbf{The need for standardized evaluation.} Our decomposition in Section~\ref{sec:benchmark} demonstrates that the six methodological differences between EEG-Bench and EEG-FM-Bench interact in ways that prevent attribution of ranking differences to model properties. We recommend the community converge on: (i) strict subject-level splits throughout; (ii) fixed checkpoint selection policies (e.g., best validation or last, but uniformly applied); (iii) standardized segment lengths per task; (iv) unified preprocessing including normalization; (v) exhaustive head ablations reported per model rather than cherry-picked. Ideally, an openly maintained benchmark platform would allow continuous community evaluation under a fixed protocol, enabling fair longitudinal comparisons as new models emerge.

\textbf{Implicit invariance through data diversity.} Multi-source pretraining spans heterogeneous acquisition systems; reconstruction forces the encoder to disentangle neural content from acquisition artifacts---an implicit form of invariance learning that simplifies the pipeline relative to explicit augmentation or multi-stage objectives.

\textbf{Dataset quantity is a confounding variable in model comparison.} The EEG foundation model literature has trended sharply toward ever-larger pretraining corpora, with corpus size frequently cited as evidence of model ambition or expected quality. Our results challenge this framing. \model{}, trained on three source corpora, matches or outperforms REVE---trained on 92 datasets and over 60{,}000 hours---on three of four protocol--task combinations. This is not a claim that data volume is unimportant: larger corpora may still confer advantages on tasks not evaluated here, particularly those requiring exposure to rare pathological patterns. Rather, it is a claim about \emph{interpretability}: when two models differ simultaneously in architecture, pretraining objective, corpus size, and corpus composition, observed performance differences cannot be attributed to any single cause. Dataset count has become a confounding variable that inflates reported differences between models while obscuring what actually drives representation quality.

The practical consequence is two-fold. First, the community risks \emph{over-collection}: assembling larger corpora from the same EU/US repositories may yield diminishing returns once basic recording-condition diversity is covered, while the cost of data curation and licensing grows linearly. Our D1-vs-D2 ablation shows that adding 9{,}663 South Asian clinical subjects---a single diverse source---produces a 12.3\,pp gain on the hardest clinical task and competitive fine-tuning performance across benchmarks, while our D1 checkpoint already matches models pretrained on far larger EU/US corpora. Second, without pretraining dataset ablations as a first-class experimental contribution, it is impossible to distinguish ``more data'' from ``better data'' from ``better architecture'' in published comparisons. We therefore recommend that future EEG foundation model papers include controlled dataset ablations alongside architecture and objective ablations, treating corpus composition as an independent experimental variable rather than a fixed condition.

\textbf{Limitations.} We use a single model size; scaling effects remain unexplored. The multi-source data confounds geography with acquisition systems, making it impossible to attribute improvements to one factor alone. Our epilepsy--mimicker dataset is drawn from South Asian clinical centers; generalization to other demographic groups requires separate validation. The masking ratio was fixed. While we show the MLP head advantage within our model, generalizability to other reconstruction-based architectures requires further investigation.

\section{Conclusion}

Through controlled ablations with fixed architecture, we found that population heterogeneity reshapes EEG representations in ways invisible to linear probing but exposed by fine-tuning. Most importantly, we demonstrated that this effect matters most where it counts: on a clinically challenging epilepsy vs.\ mimicker differential diagnosis task that no prior foundation model has addressed, the diverse checkpoint outperforms the narrow-source baseline by 12.3 pp. We provided a systematic decomposition of six concrete methodological differences between EEG-Bench and EEG-FM-Bench that produce ranking reversals of up to 24\,pp on identical models and datasets, demonstrating that benchmark inconsistency is not merely quantitative but qualitative. We further showed that \model{}, trained on three source corpora, matches or outperforms REVE---trained on 92 datasets---on the majority of evaluated protocol--task combinations, demonstrating that dataset count is a confounding variable in model comparison and that targeted diversity can substitute for indiscriminate scale. Our model matches or outperforms larger-scale models on standard benchmarks, and the cross-protocol instability we document reinforces the urgent need for community consensus on evaluation methodology, including dataset ablations as a required experimental component. We will release the 200-subject epilepsy--mimicker dataset to catalyze progress on clinically grounded EEG foundation model evaluation. Model code and implementation are available at: \url{https://osf.io/gv4p3/overview?view_only=363cb103cff44bfc9ce916240ed403a6}.


\end{document}